\documentclass[10pt,twocolumn,letterpaper]{article}
\usepackage{cvpr}
\usepackage{cite}
\usepackage{times}
\usepackage{epsfig}
\usepackage{epstopdf}
\usepackage{graphicx}
\usepackage{mathrsfs}
\usepackage{amsfonts}
\usepackage{amsmath}
\usepackage{makecell}
\usepackage{multirow}
\usepackage{balance}
\usepackage{amssymb, mathtools}
\newcommand\norm[1]{\left\lVert#1\right\rVert}
\usepackage[pagebackref=true,breaklinks=true,letterpaper=true,colorlinks,bookmarks=false]{hyperref}
\makeatletter
\newcommand\footnoteref[1]{\protected@xdef\@thefnmark{\ref{#1}}\@footnotemark}
\cvprfinalcopy
\begin{document}
\title{Real-Time Dense Stereo Embedded in A UAV for Road Inspection}

\author{Rui Fan, Jianhao Jiao, Jie Pan, Huaiyang Huang, Shaojie Shen, Ming Liu\\
HKUST Robotics Institute\\
{\tt\small rui.fan@ieee.org}
}
\maketitle
\begin{abstract}
The condition assessment of road surfaces is essential to ensure their serviceability while still providing maximum road traffic safety. This paper presents a robust stereo vision system embedded in an unmanned aerial vehicle (UAV). The perspective view of the target image is first transformed into the reference view, and this not only improves the disparity accuracy, but also reduces the algorithm's computational complexity. The cost volumes generated from stereo matching are then filtered using a bilateral filter. The latter has been proved to be a feasible solution for the functional minimisation problem in a fully connected Markov random field model. Finally, the disparity maps are transformed by minimising an energy function with respect to the roll angle and disparity projection model. This makes the damaged road areas more distinguishable from the road surface. The proposed system is implemented on an NVIDIA Jetson TX2 GPU with CUDA for real-time purposes. It is demonstrated through experiments that the damaged road areas can be easily distinguished from the transformed disparity maps.
\end{abstract}
\section{Introduction
	\label{sec.introduction}}
The frequent detection of different types of road damage, \eg, cracks and potholes, is a critical task in road maintenance \cite{Mathavan2015}. Road condition assessment  reports allow governments to appraise long-term investment schemes and allocate limited resources for road maintenance \cite{Fan2018}. However, manual visual inspection is still the main form of road condition assessment \cite{Kim2014}. This process is, however, not only tedious, time-consuming and costly, but also dangerous for the personnel \cite{Koch2011}. Furthermore, the detection results are always subjective and qualitative because decisions entirely depend on the experience of the personnel \cite{Koch2015}. Therefore, there is an ever-increasing need to develop automated road inspection systems that can recognise and localise road damage both efficiently and objectively \cite{Mathavan2015}.

Over the past decades, various technologies, such as vibration sensing, active or passive sensing, have been used to acquire road data and help technicians in assessing the road condition \cite{Koch2012}. For example, Fox \etal \cite{Fox2017} developed a crowd-sourcing system to detect road damage by analysing accelerometer data obtained from multiple vehicles.  Although vibration sensors are cost-effective and only require a small amount of storage space, the shape of a damaged road area cannot be explicitly inferred from the vibration data \cite{Kim2014}. Furthermore, Tsai \etal \cite{Tsai2017} mounted two laser scanners on a digital inspection vehicle (DIV) to collect 3D road data for pothole detection. However, such vehicles are not widely used, because of their high equipment and long-term maintenance costs \cite{Fan2018}. 

The most commonly used passive sensors for road condition assessment include Microsoft Kinect and other types of digital cameras \cite{Wang2017}. In \cite{Jahanshahi2012}, Jahanshahi \etal utilised a Kinect to acquire depth maps, from which the damaged road areas were extracted using image segmentation algorithms.  However, Kinect sensors were initially designed for indoor use, and they do not perform well when exposed to direct sunlight, causing depth values to be recorded as zero \cite{Cruz2012}. Therefore, it is more effective to detect road damages using digital cameras, as they are cost-effective and capable of working in outdoor environments \cite{Fan2018}.  

With recent advances in airborne technology, unmanned aerial vehicles (UAVs) equipped with digital cameras provide new opportunities for road inspection \cite{Schnebele2015}. For example, Feng \etal \cite{Feng2009} mounted a camera on a UAV to capture road images. The latter was then analysed to illustrate conditions such as traffic congestion, road accidents, among others.  
Furthermore, Zhang \cite{Zhang2008} designed a robust photogrammetric mapping system for UAVs, which can recognise different road defects, such as ruts and potholes, from the captured RGB images. Although the aforementioned 2D computer vision methods can recognise damaged road areas with low computational complexity,  the achieved level of accuracy is still far from satisfactory \cite{Jahanshahi2012, Koch2011}. Additionally, the structure of a detected road damage is not obvious from only a single video frame, and the depth/disparity information is more effective than RGB information in terms of detecting severe road damages, \eg, potholes \cite{Mathavan2015}. Therefore, it becomes increasingly important to use digital cameras for 3D road data acquisition. 

To reconstruct 3D road scenery using digital cameras, multiple camera views are required \cite{Hartley2003}. Images from different viewpoints can be captured using either a single movable camera or an array of synchronised cameras \cite{Fan2018}. In \cite{Zhang2012b}, Zhang and Elaksher reconstructed the 3D road scenery using structure from motion (SfM), where the keypoints in each frame were extracted using scale-invariant feature transform (SIFT) \cite{Lowe2004}, and an energy function with respect to all camera poses was optimised for accurate 3D road scenery reconstruction. However, SfM can only acquire sparse point clouds, which are usually infeasible for road damage detection \cite{Jahanshahi2012}. In this regard, many researchers have resorted to using stereo vision technology to acquire dense point clouds for road damage detection. In \cite{Fan2018}, Fan \etal developed an accurate dense stereo vision algorithm for road surface 3D reconstruction, and an accuracy of approximately $\pm$ 3 mm was achieved. However, the search range propagation strategy in their algorithm makes it difficult to fully exploit the parallel computing architecture of the graphics cards \cite{Fan2018}. Therefore, the motivation of this paper is to explore a highly efficient dense stereo vision algorithm, which can be embedded in UAVs for real-time road inspection. 


The remainder of this paper is organised as follows. Section \ref{sec.related_work} discusses the related work on stereo vision. Section \ref{sec.system_description} presents the proposed embedded stereo vision system. The experimental results for performance evaluation are provided in Section \ref{sec.experimental_results}. Finally, Section \ref{sec.conclusion} summarises the paper and provides recommendations for future work. 

\section{Related Work\label{sec.related_work}}
The two key aspects of computer stereo vision are speed and accuracy \cite{Tippetts2016a}. A lot of research has been carried out over the past decades to improve either the disparity accuracy or the algorithm's computational complexity \cite{Fan2018}. The state-of-the-art stereo vision algorithms can be  classified as convolutional neural network (CNN)-based \cite{Luo2016, Zagoruyko2015, Zbontar2015, Chang2018, Zhou2017} and traditional \cite{Fan2018, Ihler2005, Tappen2003, Boykov2001, Hirschmuller2008, Mozerov2015}. The former generally formulates  disparity estimation as a binary classification problem and learns the probability distribution over all disparity values \cite{Luo2016}. For example, PSMNet \cite{Chang2018} generates the cost volumes by learning region-level features with different scales of receptive fields. Although these approaches have achieved some highly accurate disparity maps, they usually require a large amount of labelled training data to learn from. Therefore, it is impossible for them to work on the datasets without providing the disparity ground truth \cite{Zhou2017}. Moreover, predicting disparities with CNNs is still a computationally intensive task, which usually takes seconds or even minutes to execute on state-of-the-art graphics cards \cite{Tippetts2016a}. Therefore, the existing CNN-based stereo vision algorithms are not suitable for real-time applications. 

The traditional stereo vision algorithms can be classified as local, global and semi-global \cite{Fan2018}. The local algorithms typically select a series of blocks from the target image and match them with a constant block selected from the reference image \cite{Fan2018}. The disparities are then determined by finding the shifting distances corresponding to either the highest correlation or the lowest cost \cite{Tippetts2016a}. This optimisation technique is also known as winner-take-all (WTA).  

Unlike the local algorithms, the global algorithms generally translate stereo matching into an energy minimisation problem, which can later be addressed using sophisticated optimisation techniques, \eg, belief propagation (BP) \cite{Ihler2005} and graph cuts (GC) \cite{Boykov2001}. These techniques are commonly developed based on the Markov random field (MRF) \cite{Tappen2003}. Semi-global matching (SGM) \cite{Hirschmuller2008} approximates the MRF inference by performing cost aggregation along all directions in the image, and this greatly improves the accuracy and efficiency of stereo matching. However, finding the optimum smoothness values is a challenging task, due to the occlusion problem \cite{Mozerov2015}. Over-penalising the smoothness term can reduce ambiguities around the discontinuous areas, but on the other hand, can cause incorrect matches for the continuous areas \cite{Fan2018}. Furthermore, the computational complexities of the aforementioned optimisation techniques are significantly intensive, making these algorithms difficult to perform in real time \cite{Tippetts2016a}. 

In \cite{Fan2018}, Fan \etal proposed a novel perspective transformation method, which improves both the disparity accuracy and the computational complexity of the algorithm. Furthermore, Mozerov and Weijer \cite{Mozerov2015} proved that bilateral filtering is a feasible solution for the energy minimisation problem in a fully connected MRF model. The costs can be adaptively aggregated by performing bilateral filtering on the initial cost volumes \cite{Fan2018}. Therefore, the proposed stereo vision system is developed based on the work in \cite{Fan2018} and \cite{Mozerov2015}. Finally, the estimated disparity maps are transformed by minimising an energy function with respect to the roll angle and disparity projection model. This makes the damaged road areas become highly distinguishable from the road surface.  
\section{System Description\label{sec.system_description}}
\begin{figure*}[!t]
	\begin{center}
		\centering
		\includegraphics[width=0.999\textwidth]{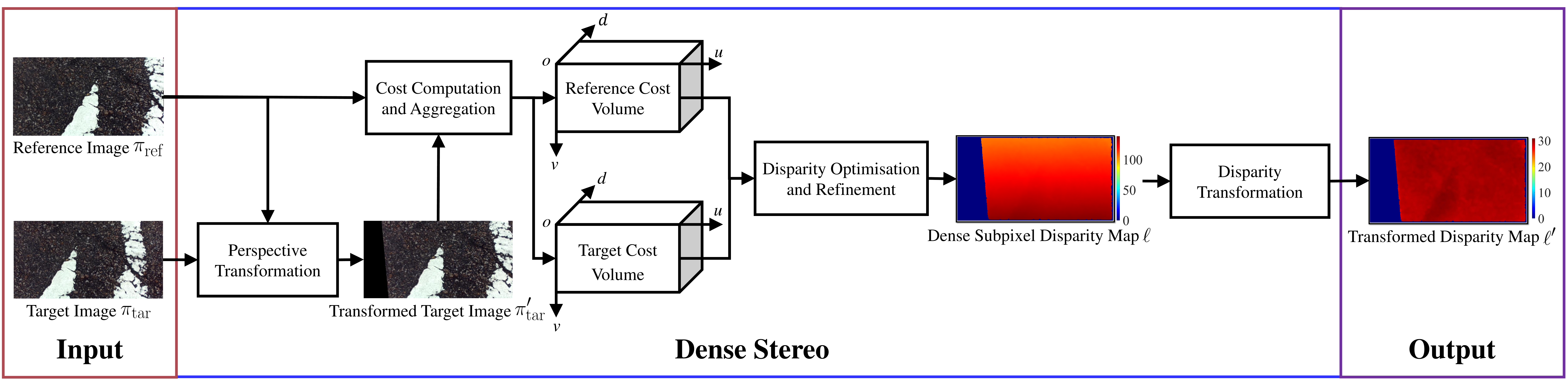}
		\caption{Workflow of the proposed dense stereo system. }
		\label{Figure workflow}
				\vspace{-1em}
	\end{center}
\end{figure*}

The workflow of the proposed stereo vision system is depicted in Figure  \ref{Figure workflow}, where the system consists of three main components: a) perspective transformation; b) dense road stereo; and c) disparity transformation.  The following subsections describe each component in turn. 
\subsection{Perspective Transformation\label{sec.perspective_transformation}}
In this paper, the road surface is treated as a ground plane:
\begin{equation}
\mathbf{n}^\top \mathbf{p^\text{W}}+\beta=0,
\label{eq.road_surface_func}
\end{equation}
where $\mathbf{p^\text{W}}=[x^\text{W},y^\text{W},z^\text{W}]^\top$ is a 3D point on the road surface in the world coordinate system (WCS), and $\mathbf{n}=[n_x, n_y, n_z]^\top$ is the normal vector of the ground plane. The projections of $\mathbf{p^\text{W}}$ on the reference and target images, \ie, $\pi_\text{ref}$ and $\pi_\text{tar}$, are  $\mathbf{p}^\text{I}_\text{ref}=[u_\text{ref},v_\text{ref}]^\top$ and $\mathbf{p}^\text{I}_\text{tar}=[u_\text{tar},v_\text{tar}]^\top$, respectively. It should be noted that the left and right images are respectively referred to as the reference and target images in this paper.  $\mathbf{p}^\text{I}_\text{ref}$ can be transformed to  $\mathbf{p}^\text{I}_\text{tar}$ using a homography matrix $\mathbf{H}$ as follows \cite{Fan2018}:
\begin{equation}
\begin{split}
\begin{bmatrix}
{\mathbf{p}^\text{I}_\text{tar}}\\1
\end{bmatrix}
=\mathbf{H}\begin{bmatrix}
{\mathbf{p}^\text{I}_\text{ref}}\\1
\end{bmatrix},
\end{split}
\label{eq.homograph_mat}
\end{equation}
where 
\begin{equation}
\mathbf{H}=
\mathbf{K}_\text{tar}
\left(\mathbf{R}_1 -\frac{\mathbf{t}\mathbf{n}^\top{\mathbf{R}_0}^{-1}}{\beta}
\right)
{\mathbf{K}_\text{ref}}^{-1},
\label{eq.H_dec}
\end{equation}
$\mathbf{t}$ is a translation vector, $\mathbf{R}_0$ represents the rotation from the WCS to the reference camera coordinate system (RCCS), $\mathbf{R}_1$ denotes the rotation from the RCCS to the target camera coordinate system (TCCS), and $\mathbf{K}_\text{ref}$ and $\mathbf{K}_\text{tar}$ are the intrinsic matrices of the reference and target cameras, respectively. $\mathbf{H}$ can be estimated using at least four pairs of matched correspondence points $\mathbf{p}^\text{I}_\text{ref}$ and $\mathbf{p}^\text{I}_\text{tar}$ \cite{Hartley2003}. In order to simplify the estimation of $\mathbf{H}$, the authors of \cite{Fan2018} made several hypotheses regarding $\mathbf{R}_0$, $\mathbf{R}_1$, $\mathbf{K}_\text{ref}$, $\mathbf{K}_\text{tar}$, $\mathbf{t}$ and $\mathbf{n}$.  (\ref{eq.homograph_mat})  can be rewritten as follows:
\begin{equation}
u_\text{tar}=u_\text{ref}+\frac{t_cn_x}{\beta}(f\sin\theta-v_o\cos\theta)+v\frac{t_cn_x}{\beta}\cos\theta,
\label{eq.relation_vd}
\end{equation}
where $f$ is the focus length of each camera, $t_c$ is the baseline, $\theta$ is the pitch angle, and $[u_o,v_o]^\top$ is the principal point. $v=v_\text{ref}=v_\text{tar}$.  (\ref{eq.relation_vd}) implies that a perspective distortion always exists for the ground plane in two images when $\theta$ is not equal to $\pi/2$, and this further affects the  stereo matching accuracy. Therefore, the perspective transformation aims to make the ground plane in the transformed target image similar to that in the reference image \cite{Fan2018}. This can be straightforwardly realised by shifting each point on row $v$ in the target image $\Delta u-\delta_p$ pixels to the right, where $\Delta u=u_\text{ref}-u_\text{tar}$, and $\delta_p$ is a constant used to guarantee that all the disparities are non-negative. The values of $t_c$, $n_x$, $f$, $\beta$, $v_o$ and $\theta$ can be estimated from a set of reliable correspondence pairs $\mathbf{Q}_\text{ref}=[\mathbf{{p}}^\text{I}_{{\text{ref}}_{0}},\mathbf{{p}}^\text{I}_{{\text{ref}}_{1}},\dots,\mathbf{{p}}^\text{I}_{{\text{ref}}_{n}}]^\top$ and $\mathbf{Q}_\text{tar}=[\mathbf{{p}}^\text{I}_{{\text{tar}}_{0}},\mathbf{{p}}^\text{I}_{{\text{tar}}_{1}},\dots,\mathbf{{p}}^\text{I}_{{\text{tar}}_{n}}]^\top$. The transformed target image is shown in Figure  \ref{Figure workflow} as $\pi_\text{tar}'$. 
\subsection{Dense Road Stereo\label{sec.dense_stereo}}
\subsubsection{Cost Computation and Aggregation\label{sec.cost_computation_aggregation}}
According to \cite{Mozerov2015}, finding the best disparities is equivalent to maximising the joint probability in (\ref{eq.mrf_eq1}):
\begin{equation}
P(\mathbf{p}_{ij}, q)=\prod_{\mathbf{p}_{ij}\in\mathscr{P}} \Phi(\mathbf{p}_{ij}, q_{\mathbf{p}_{ij}})\prod_{\mathbf{n}_{\mathbf{p}_{ij}}\in\mathscr{N}_{\mathbf{p}_{ij}}} \Psi (\mathbf{p}_{ij}, \mathbf{n}_{\mathbf{p}_{ij}}),
\label{eq.mrf_eq1}
\end{equation}
where $\mathbf{p}_{ij}$ denotes a node at the position of $(i,j)$ in the graph $\mathscr{P}$,  $q_{\mathbf{p_{ij}}}$ represents the intensity differences corresponding to different disparities $d$, 
$\mathscr{N}_{\mathbf{p}_{ij}}=\{\mathbf{n}_{{\mathbf{p}_{ij}}_1},\mathbf{n}_{{\mathbf{p}_{ij}}_2},\mathbf{n}_{{\mathbf{p}_{ij}}_3},\cdots,\mathbf{n}_{{\mathbf{p}_{ij}}_k}|\mathbf{n}_{\mathbf{p}_{ij}}\in\mathscr{P}\}$ represents the neighbourhood system {of} $\mathbf{p}_{ij}$,  $\Phi(\cdot)$ expresses the compatibility between  each possible disparity $d$ and the corresponding intensity difference, and $\Psi(\cdot)$ expresses the compatibility between $\mathbf{p}_{ij}$ and its neighbourhood system $\mathscr{N}_{\mathbf{p}_{ij}}$.  It is noteworthy that $\mathbf{p}_{uv}$ refers to $\mathbf{p}^\text{I}_\text{ref}=[u_\text{ref}, v_\text{ref}]^\top$ and $\mathscr{P}$ refers to the reference image. In practice, maximising the joint probability in (\ref{eq.mrf_eq1}) is commonly formulated as an energy minimisation problem as follows \cite{Fan2018c}:
\begin{equation}
E_d(\mathbf{p}_{ij},d)=\sum_{\mathbf{p}_{ij}\in\mathscr{P}} D(\mathbf{p}_{ij},d)+
\sum_{\mathbf{n}_{\mathbf{p}_{ij}}\in\mathscr{N}_{\mathbf{p}_{ij}}} V (\mathbf{p}_{ij}, \mathbf{n}_{\mathbf{p}_{ij}},d),
\label{eq.mrf_eq2}   
\end{equation}
where $D(\cdot)$ computes the matching cost of $\mathbf{p}_{ij}$, and $V(\cdot)$ determines the aggregation strategy. For disparity estimation algorithms based on the MRF, formulating $V(\cdot)$ in an adaptive way is crucial and necessary, because the intensity of a pixel in a discontinuous area usually differs greatly from those of its neighbours \cite{Mozerov2015}. Since bilateral filtering is a feasible solution for the energy minimisation problem in a fully connected MRF model \cite{Mozerov2015}, $D(\cdot)$ and $V(\cdot)$ can be rewritten as follows:
\begin{equation}
D(\mathbf{p}_{ij},d)=c(\mathbf{p}_{ij},d),
\label{eq.D}   
\end{equation}
where
\begin{equation}
\begin{split}
c(\mathbf{p},d)&=
\frac{(\sigma_\text{ref}\sigma_\text{tar}+\mu_\text{ref}\mu_\text{tar})}{\sigma_\text{ref}\sigma_\text{tar}}
\\-&\frac{1}{n\sigma_\text{ref}\sigma_\text{tar}}\Bigg(
\sum_{\mathbf{q}\in\mathscr{N}^{+}_{\mathbf{p}}}i_\text{ref}(\mathbf{q})i_\text{tar}(\mathbf{q}-[d,0]^\top)
\Bigg)
\end{split}
\label{eq.c}   
\end{equation}
is the cost function; $i_\text{ref}(\mathbf{p})$ and $i_\text{tar}(\mathbf{p})$ represent the pixel intensities at $\mathbf{p}$ in the reference and target images, respectively; $\mu_\text{ref}$ and $\mu_\text{tar}$ represent the means of the pixel intensities within the reference and target blocks, respectively; and $\sigma_\text{ref}$ and $\sigma_\text{tar}$ denote the standard deviations of the reference and target blocks, respectively. $\mathscr{N}_{\mathbf{p}}^+=\{\mathbf{p}\}\cup\mathscr{N}_{\mathbf{p}}$.
\begin{equation}
V(\mathbf{p}_{ij},\mathbf{n}_{\mathbf{p}_{ij}},d)=\sum_{\mathbf{n}_{\mathbf{p}_{ij}}\in\mathscr{N}_{\mathbf{p}_{ij}}}\omega(\mathbf{p}_{ij}, \mathbf{n}_{\mathbf{p}_{ij}}) c(\mathbf{n}_{\mathbf{p}_{ij}},d),
\label{eq.V}   
\end{equation}
where 
\begin{equation}
\omega(\mathbf{p}, \mathbf{n}_{\mathbf{p}})=\exp \bigg\{ -\frac{\norm{\mathbf{p}-\mathbf{n}_{\mathbf{p}}}_2^2}{{\sigma_0}^2} -\frac{(i_\text{ref}(\mathbf{p})-i_\text{ref}(\mathbf{n}_{\mathbf{p}}))^2}{{\sigma_1}^2} \bigg\}
\label{eq.omega_s}
\end{equation}
is controlled by two parameters $\sigma_0$ and $\sigma_1$, with $\sigma_0$ based on spatial distance and  $\sigma_1$ based on colour similarity.  The cost $c$ of each neighbour $\mathbf{n}_{\mathbf{p}}$ can therefore be adaptively aggregated to $\mathbf{p}$. Finally, $E_d(\mathbf{p},d)$ is normalised by rewriting (\ref{eq.mrf_eq2}) as follows:
\begin{equation}
E_d(\mathbf{p},d)=\frac{\sum\limits_{{\mathbf{q}}\in\mathscr{N}_{\mathbf{p}}^+} \omega(\mathbf{p},\mathbf{q})D(\mathbf{q},d)}{\sum\limits_{{\mathbf{q}}\in\mathscr{N}_{\mathbf{p}}^+} \omega(\mathbf{p},\mathbf{q})},
\label{eq.fbs}
\end{equation}
The computed matching costs are stored in two cost volumes, as shown in Figure  \ref{Figure workflow}. 
\subsubsection{Disparity Optimisation and Refinement\label{sec.disparity_optimisation_refinement}}
By applying WTA optimisation on the reference and target cost volumes, the best disparities can be estimated. Since the perspective view of the target image has been transformed in Section \ref{sec.perspective_transformation}, the estimated disparities on row $v$ should be added $\Delta u-\delta_p$ to obtain the disparity map between the original reference and target images. The occluded areas in the reference disparity map are then removed by finding the pixels $\mathbf{p}$ satisfying the following condition \cite{Fan2018b}:
\begin{equation}
\norm{{\ell_\text{ref}(\mathbf{p})-\ell_\text{tar}(\mathbf{p}- [\ell_\text{ref}(\mathbf{p}),0]^\top})}^2_2>\delta_r,
\label{eq.lrc}
\end{equation}
where $\ell_\text{ref}$ and $\ell_\text{tar}$ represent the reference and target disparity maps, respectively. $\delta_r=1$ is the threshold for occlusion removal. Finally, a subpixel enhancement is performed to increase the resolution of the estimated disparity values \cite{Fan2018}: 
\begin{equation}
\ell(\mathbf{p})=\ell_\text{ref}(\mathbf{p})+\frac{c(\mathbf{p},d-1)-c(\mathbf{p},d+1)}{2c(\mathbf{p},d-1)+2c(\mathbf{p},d+1)-4c(\mathbf{p},d)},
\label{eq.subpixel}
\end{equation}
where $\ell$, illustrated in Figure  \ref{Figure workflow}, represents the final disparity map in the reference perspective view. 
\subsection{Disparity Transformation\label{disparity_transformation}}
The proposed system focuses entirely on the road surface whose disparity values decrease gradually from the bottom of the disparity map to its top, as shown in Figure  \ref{Figure workflow}. For a stereo rig whose baseline is perfectly parallel to the road surface, the roll angle $\psi$ equals zero and the disparities on each row have similar values, which can also be proved by (\ref{eq.relation_vd}). Therefore, the projection of the road disparities on a v-disparity image can be represented by a linear model: $f(v)=\alpha_0+\alpha_1 v$. A column vector $\boldsymbol{\alpha}=[\alpha_0,\alpha_1]^\top$ storing the coefficients of the disparity projection model can be estimated as follows:
\begin{equation}
\boldsymbol{\alpha}=\underset{\boldsymbol{\alpha}}{\arg\min}\ E_t,
\label{eq.initial_energy_minimisation}
\end{equation}
where
\begin{equation}
E_t=\norm{\mathbf{d}-\mathbf{V}\boldsymbol{\alpha}}_2^2,
\label{eq.E}
\end{equation}
$\mathbf{d}=[\ell(\mathbf{p}_0),\  \ell(\mathbf{p}_1),\ \cdots,\  \ell(\mathbf{p}_n)]^\top$ stores the disparity values, $\mathbf{v}=[v_0,\  v_1,\  \cdots,\  v_n]^\top$ stores the vertical disparity coordinates, $\mathbf{1}_k$ represents a $k\times1$ vector of ones, and $\mathbf{V}=[\mathbf{1}_{n+1} \  \mathbf{v}]$. Applying  (\ref{eq.E}) to (\ref{eq.initial_energy_minimisation}) results in the following expression:
\begin{equation}    
\boldsymbol{\alpha}=(\mathbf{V}^\top\mathbf{V})^{-1}\mathbf{V}^{\top}\mathbf{d}.
\label{eq.alpha2}
\end{equation}
The minimum energy ${E_t}_{\text{min}}$  can be obtained by applying (\ref{eq.alpha2}) to  (\ref{eq.E}):
\begin{equation}
\begin{split}
{E_t}_\text{min}
=\mathbf{d}^\top\mathbf{d}-\mathbf{d}^\top\mathbf{V}(\mathbf{V}^\top\mathbf{V})^{-1}\mathbf{V}^\top\mathbf{d}.
\end{split}
\label{eq.E_min}
\end{equation}
However, in practice, the stereo rig baseline is not always perfectly parallel to the road surface, and this introduces a non-zero roll angle $\psi$ into the imaging process.  The disparity values will change gradually in the horizontal direction, and this makes the approach of representing the road disparity projection using a linear model problematic. Additionally, the minimum energy ${E_t}_\text{min}$ becomes higher, due to the disparity dispersion in the horizontal direction. Hence, the proposed disparity transformation first finds the angle corresponding to the minimum ${E_t}_\text{min}$. The image rotation caused by $\psi$ is then eliminated, and $\boldsymbol{\alpha}$ is subsequently estimated. 

To rotate the disparity map around a given angle $\psi$, each set of original coordinates $[u,v]^\top$ is transformed to a set of new coordinates $[x(\psi), y(\psi)]^\top$ using the following equations \cite{Fan2018a}:
\begin{equation}
x(\psi)=u\cos\psi+v\sin\psi,
\label{eq.x}
\end{equation}
\begin{equation}
y(\psi)=v\cos\psi-u\sin\psi.
\label{eq.y}
\end{equation}
The energy function in (\ref{eq.E}) can, therefore, be rewritten as follows:
\begin{equation}
E_t(\psi)=\norm{\mathbf{d}-\mathbf{Y}(\psi)\boldsymbol{\alpha}}_2^2,
\label{eq.E2}
\end{equation}
where $\mathbf{y}=[y_0(\psi),\ y_1(\psi),\ \cdots,\ y_n(\psi)]^\top$ and $\mathbf{Y}(\psi)=[\mathbf{1}_{n+1} \  \mathbf{y}(\psi)]$. (\ref{eq.alpha3}) is obtained by applying (\ref{eq.E2}) to (\ref{eq.initial_energy_minimisation}):
\begin{equation}
\boldsymbol{\alpha}(\psi)=\mathbf{J}(\psi)\mathbf{d},
\label{eq.alpha3}
\end{equation}
where
\begin{equation}
\mathbf{J}(\psi)=(\mathbf{Y}(\psi)^\top\mathbf{Y}(\psi))^{-1}\mathbf{Y}(\psi)^{\top}.
\label{eq.J}
\end{equation}
${E_t}_\text{min}$ can also be obtained by applying (\ref{eq.alpha3}) and (\ref{eq.J}) to (\ref{eq.E2}):
\begin{equation}
{{E}_t}_\text{min}(\psi)=\mathbf{d}^\top\mathbf{d}-\mathbf{d}^\top\mathbf{Y}\big(\mathbf{Y}(\psi)^\top\mathbf{Y}(\psi)\big)^{-1}\mathbf{Y}(\psi)^\top\mathbf{d}.
\label{eq.E_min2}
\end{equation}
Roll angle estimation  is, therefore, equivalent to the following energy minimisation problem:
\begin{equation}
\psi=\underset{\psi}{\arg\min}\ {E_t}_\text{min}(\psi) \ \ \text{s.t.} \ \psi\in(-\frac{\pi}{2},\frac{\pi}{2}], 
\label{eq.e_min_min}
\end{equation}
which can be formulated as an iterative optimisation problem as follows \cite{Pedregal2006}:
\begin{equation}
\psi^{(k+1)}=\psi^{(k)}-\lambda{\nabla{E}_t}_\text{min}({\psi}^{(k)}),\ \ \ k\in\mathbb{N}^0,
\label{eq.e_min_gd}
\end{equation}
where $\lambda$ is the learning rate. (\ref{eq.e_min_gd}) is a standard form of gradient descent. The expression of  $\nabla {E_t}_\text{min}$ is as follows:
\begin{equation}
\nabla {E_t}_\text{min}(\psi)=-2\mathbf{d}^\top\mathbf{W}(\psi)\mathbf{d}, 
\label{eq.d_e_min}
\end{equation}
where
\begin{equation}
\begin{split}
\mathbf{W}(\psi)=\Big(\mathbf{I}-\mathbf{Y}(\psi)\mathbf{J}(\psi)\Big)\nabla\mathbf{Y}(\psi)\mathbf{J}(\psi),
\end{split}
\label{eq.M}
\end{equation}
$\mathbf{I}$ is an identity matrix. If $\lambda$ is too high, (\ref{eq.e_min_gd}) may overshoot the minimum. On the other hand, if $\lambda$ is set to a relatively low value, the convergence of (\ref{eq.e_min_gd}) may require a lot of  iterations \cite{Pedregal2006}. Therefore, selecting a proper $\lambda$ is always essential for gradient descent. Instead of fixing the learning rate with a constant value, backtracking line search is utilised to produce an adaptive learning rate: 
\begin{equation}
\lambda^{(k+1)}=\frac{\lambda^{(k)}\nabla{{E}_t}_\text{min}({\psi}^{(k)})}{\nabla{{E}_t}_\text{min}({\psi}^{(k)})-\nabla{{E}_t}_\text{min}({\psi}^{(k+1)})},\ \ \  k\in\mathbb{N}^0.
\label{eq.adaptive_lambda}
\end{equation}
The selection of the initial learning rate $\lambda^{(0)}$ will be discussed in Section \ref{sec.experimental_results}. The initial approximation $\psi^{(0)}$ is set to $0$, because the roll angle in practical experiments is usually small. It should be noted that the estimated $\psi$ at time $t$ is used as the initial approximation at time $t+1$. The optimisation iterates until the absolute difference between $\psi^{(k)}$ and $\psi^{(k+1)}$ is smaller than a preset threshold $\delta_\psi$. $\boldsymbol{\alpha}$ can be obtained by substituting the estimated roll angle $\psi$ into (\ref{eq.alpha3}). Finally, each disparity is transformed using:
\begin{equation}
{\ell}'(\mathbf{p})={\ell}(\mathbf{p})-\alpha_0+\alpha_1(u\sin\psi-v\cos\psi)+\delta_t,
\label{eq.disparity_transformation}
\end{equation}
where ${\ell}'$, shown in Figure  \ref{Figure workflow}, represents the transformed disparity map, and $\delta_t$ is a constant used to make the transformed disparity values positive.

\section{Experimental Results\label{sec.experimental_results}}
In this section, we evaluate the performance of the proposed stereo vision system both qualitatively and quantitatively. The following subsections detail the experimental set-up, datasets, implementation notes and the performance evaluation.
\subsection{Experimental Set-Up\label{sec.experimental_setup}}
\begin{figure}[!t]
	\begin{center}
		\centering
		\includegraphics[width=0.28\textwidth]{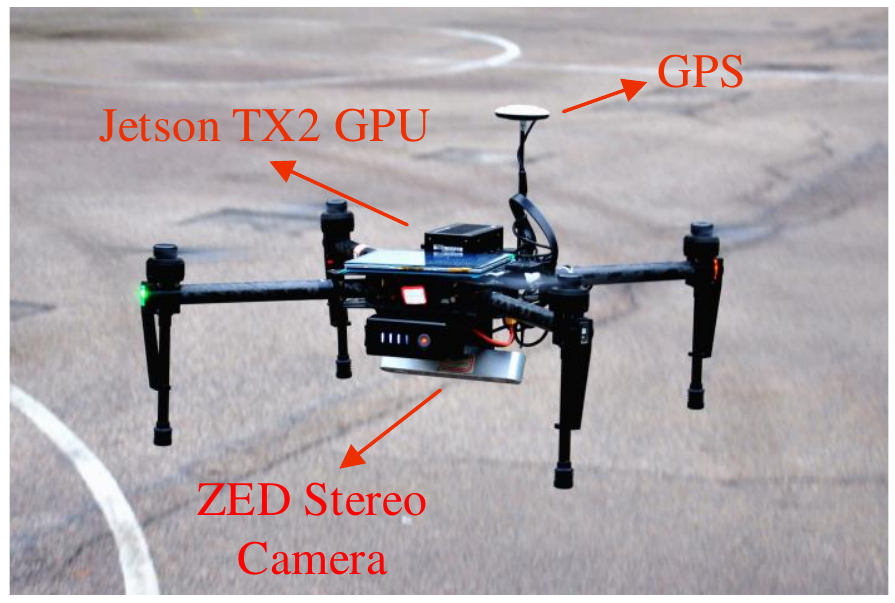}
		\caption{Experimental set-up. }
		\vspace{-1em}
		\label{Figure experimental_set_up}
	\end{center}
\end{figure}
\begin{figure*}[!t]
	\begin{center}
		\centering
		\includegraphics[width=0.95\textwidth]{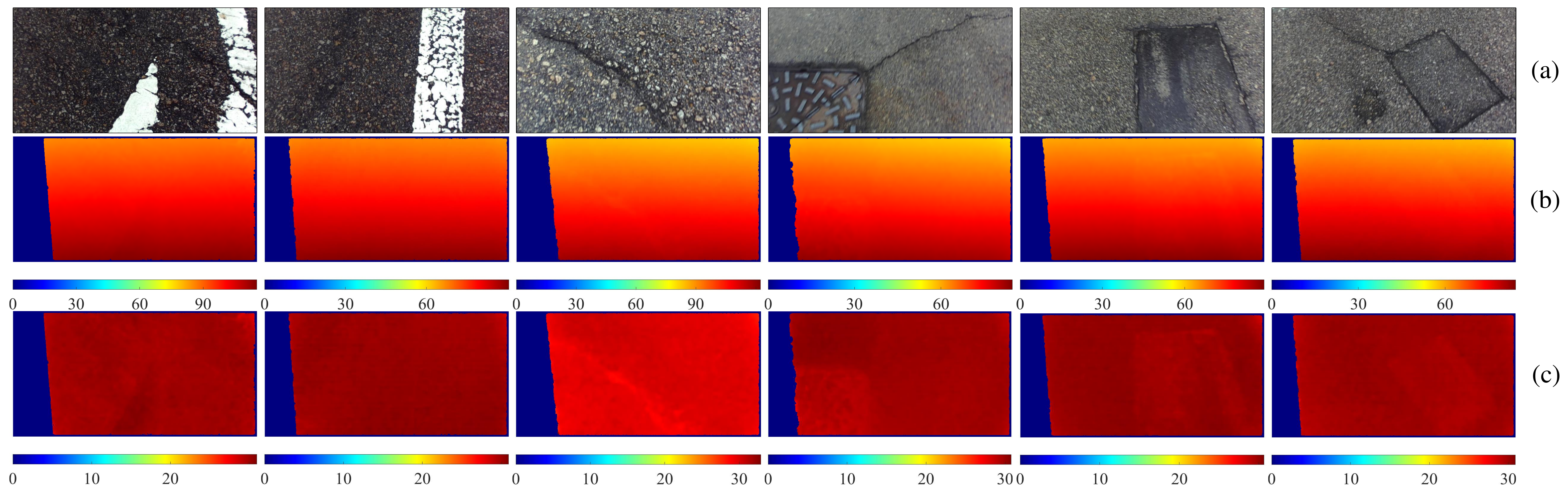}
		\caption{Experimental results; (a) reference  images; (b) dense subpixel disparity maps; (c) transformed disparity maps. }
				\vspace{-1.1em}
		\label{Figure results}
	\end{center}
\end{figure*}
\begin{figure*}[!t]
	\begin{center}
		\centering
		\includegraphics[width=0.94\textwidth]{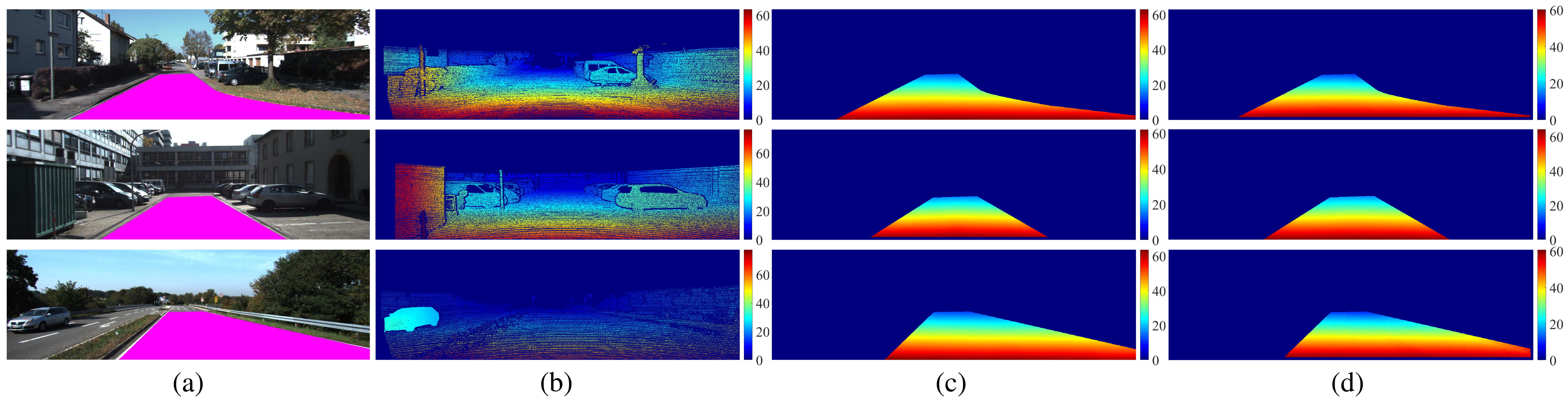}
		\caption{Examples of the KITTI stereo experimental results; (a) reference images, where the areas in magenta are the manually selected road regions; (b) ground truth disparity maps; (c) results obtained using PSMNet; (d) results obtained using the proposed algorithm. }
				\vspace{-1.1em}
		\label{Figure kitti_disparity}
	\end{center}
\end{figure*}
In the experiments, a ZED stereo camera\footnote{https://www.stereolabs.com/\label{note.zed}} is mounted on a DJI Matrice 100 Drone\footnote{https://www.dji.com/uk/matrice100\label{note.dji}} to capture stereo road images. The maximum take-off weight of the drone is 3.6 kg. The stereo camera has two ultra-sharp six-element all-glass lenses, which can cover the scene up to 20 m\footnoteref{note.zed}. The captured stereo road images are processed using an NVIDIA Jetson TX2 GPU\footnote{https://developer.nvidia.com/embedded/buy/jetson-tx2\label{note1.jetson}}, which has 8 GB LPDDR4 memory and 256 CUDA cores.  An illustration of the experimental set-up is shown in Figure  \ref{Figure experimental_set_up}.
\subsection{Datasets \label{sec.dataset}}
Using the above experimental set-up, three datasets including 11368 stereo image pairs are created. The resolution of the original reference and target images is $640\times360$. In each dataset, the UAV flight trajectory forms a closed loop, which makes it possible to evaluate the performance of the state-of-the-art visual odometry algorithms using our created datasets. The datasets and a demo video are publicly available at \url{http://www.ruirangerfan.com}.
\subsection{Implementation Notes\label{sec.implementation_notes}}
In the practical implementation, the reference and target images are first sent to the global memory of the GPU from the host memory. However, a thread is more likely to fetch the data from the closest addresses that its nearby threads accessed\footnote{https://docs.nvidia.com/cuda/pdf/CUDA\_C\_Programming\_Guide.pdf\label{note.gpu}}. This fact makes the use of cache in global memory impossible. Furthermore, constant memory and texture memory are read-only and cached on-chip, and this makes them more efficient than global memory for memory requesting\footnoteref{note.gpu}. Therefore, we store the reference and target images in the texture memory to reduce the memory requests from the global memory. This is realised by creating two texture objects in the texture memory and binding these objects with the addresses of the reference and target images. The pixel intensities can therefore be fetched from the texture objects instead of the global memory. In addition, (\ref{eq.omega_s}) is rewritten as follows:
\begin{equation}
\omega(\mathbf{p}, \mathbf{n}_{\mathbf{p}})=\omega_0(\mathbf{p}, \mathbf{n}_{\mathbf{p}})\omega_1(\mathbf{p}, \mathbf{n}_{\mathbf{p}}),
\label{eq.omega}
\end{equation}
where
\begin{equation}
\omega_0(\mathbf{p}, \mathbf{n}_{\mathbf{p}})=\exp \bigg\{ -\frac{\norm{\mathbf{p}-\mathbf{n}_{\mathbf{p}}}_2^2}{{\sigma_0}^2} \bigg\}
\label{eq.omega0}
\end{equation}
and
\begin{equation}
\omega_1(\mathbf{p}, \mathbf{n}_{\mathbf{p}})=\exp \bigg\{ -\frac{(i_\text{ref}(\mathbf{p})-i_\text{ref}(\mathbf{n}_{\mathbf{p}}))^2}{{\sigma_1}^2} \bigg\}.
\label{eq.omega1}
\end{equation}
The values of $\omega_0$ and $\omega_1$ are pre-calculated and stored in the constant memory to reduce the repetitive computations of $\omega$. Moreover, the values of $\mu_\text{ref}$, $\mu_\text{tar}$, $\sigma_\text{ref}$ and $\sigma_\text{tar}$ are also pre-calculated and stored in the global memory to avoid the unnecessary computations in stereo matching. 
\subsection{Performance Evaluation\label{sec.performance_evaluation}}
\subsubsection{Disparity Estimation}

Some experimental results are illustrated in Figure  \ref{Figure results}. $\mathscr{N}$ is a 120-connected neighbourhood system. $\sigma_0$ and $\sigma_1$ are empirically set to $1.5$ and $5.5$, respectively.
Since the datasets we created do not contain disparity ground truth, the KITTI\footnote{http://www.cvlibs.net/datasets/kitti/} stereo 2012 and 2015 datasets \cite{Geiger2012, Menze2015} are utilised to quantify the accuracy of the proposed system. Some experimental results of the KITTI stereo datasets are shown in Figure  \ref{Figure kitti_disparity}, where the road regions are manually selected to  evaluate the accuracy of the road disparities. 
Furthermore, we compare the proposed method with PSMNet \cite{Chang2018} in terms of the percentage of error pixels $e_p$ and root mean squared error $e_r$. The expressions of  $e_p$ and $e_r$ are as follows:
\begin{equation}
e_{p}=\frac{1}{m}\sum_{\mathbf{p}}\delta\big(  
|\ell(\mathbf{p})-\tilde{\ell}(\mathbf{p})|,
\varepsilon_{d}\big)\times100\%,
\label{eq.e_pep}
\end{equation}
\begin{equation}
e_{r}=\sqrt{\frac{1}{m}\sum_{\mathbf{p}}\big(  
	\ell(\mathbf{p})-\tilde{\ell}(\mathbf{p})
	\big)^2},
\label{eq.e_rms}
\end{equation}
where 	\begin{equation}
\delta(x, \varepsilon_d)=\bigg\{
\begin{aligned}
1 \ \ \ \ \ (x>\varepsilon_d)\\
0 \ \ \ \ \ (x\leq\varepsilon_d)
\end{aligned},
\label{eq.delta_function}
\end{equation}
$m$ is the total number of disparities used for evaluation, $\varepsilon_{d}$ is the disparity error tolerance, and $\tilde{\ell}$ represents the ground truth disparity map. 
The comparison of $e_p$ and $e_r$ between these two methods is shown in Table \ref{table.comparison}, where 
\begin{figure}[!t]
	\begin{center}
		\centering
		\includegraphics[width=0.48\textwidth]{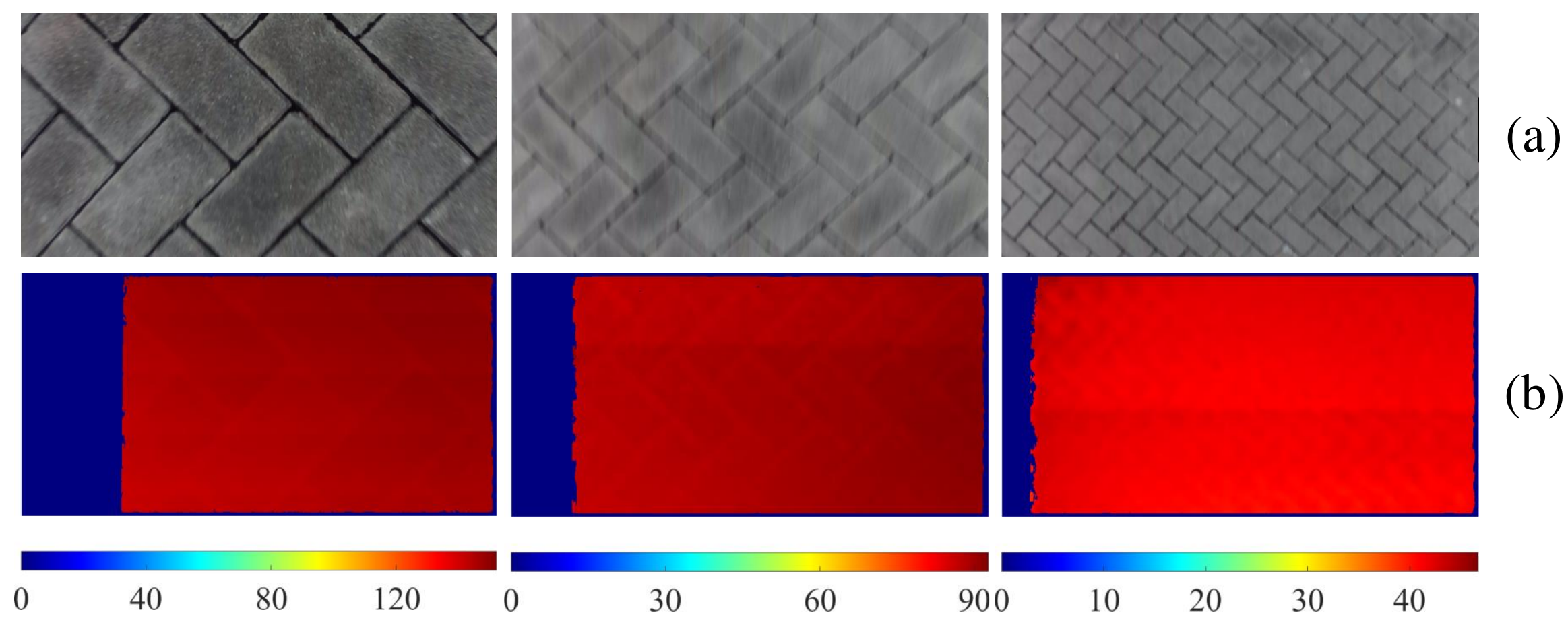}
		\caption{Disparity maps of some motion blurred images; (a) reference images; (b) disparity maps. }
				\vspace{-1em}
		\label{Figure motion_blur}
	\end{center}
\end{figure}  
it can be observed that the proposed method outperforms PSMNet in terms of $e_p$ and $e_r$ when $\varepsilon_d$ is set to 2, while PSMNet performs better than our method when $\varepsilon_d$ is set to 3. It should be noted that the proposed algorithm is capable of estimating disparity maps between a pair of motion blurred stereo images, as shown in Figure  \ref{Figure motion_blur}.  This also demonstrates the robustness of the proposed dense stereo system. 
\begin{table}[!h]
	\begin{center}
		\footnotesize
		\begin{tabular}{|c|c|cc|}
			\hline
			\multirow{2}{*}{Method} & \multirow{2}{*}{$e_r$} & \multicolumn{2}{c|}{$e_p$} \\
			\cline{3-4}
			&  & $\varepsilon_d=2$ & $\varepsilon_d=3$ \\
			\cline{3-4}	
			\hline
			\hline
			PSMNet& 1.039 & 1.345 & ${0.016}$  \\
			Ours & ${0.409}$ & ${0.217}$ & 0.023 \\
			\hline	
		\end{tabular}
	\end{center}
	\caption{Comparison between PSMNet and the proposed method in terms of disparity accuracy.}\label{table.comparison}
\end{table}

In addition to the disparity accuracy, the execution speed of the proposed dense stereo vision system is also quantified to evaluate the overall system's  performance. Owing to the fact that the image size and disparity range are not constant among different datasets, a general way of evaluating the performance in terms of processing speed is to measure  millions of disparity evaluations per second \cite{Tippetts2016a}:
\begin{equation}
Mde/s=\frac{u_\text{max}v_\text{max}d_\text{max}}{t}\times10^{-6},
\end{equation}
where the resolution of the disparity map is $u_\text{max}\times v_\text{max}$, $d_\text{max}$ is the maximum disparity value, and $t$ is the processing time in seconds.
The runtime of the proposed dense stereo vision system on the Jetson TX2 GPU is approximately 152.889 ms, and the resolution of the disparity map is $695\times361$. Therefore, the value of $Mde/s$ is 49.231, which is much higher than most stereo vision systems implemented on powerful graphics cards. 

\begin{figure}[!t]
	\begin{center}
		\centering
		\includegraphics[width=0.42\textwidth]{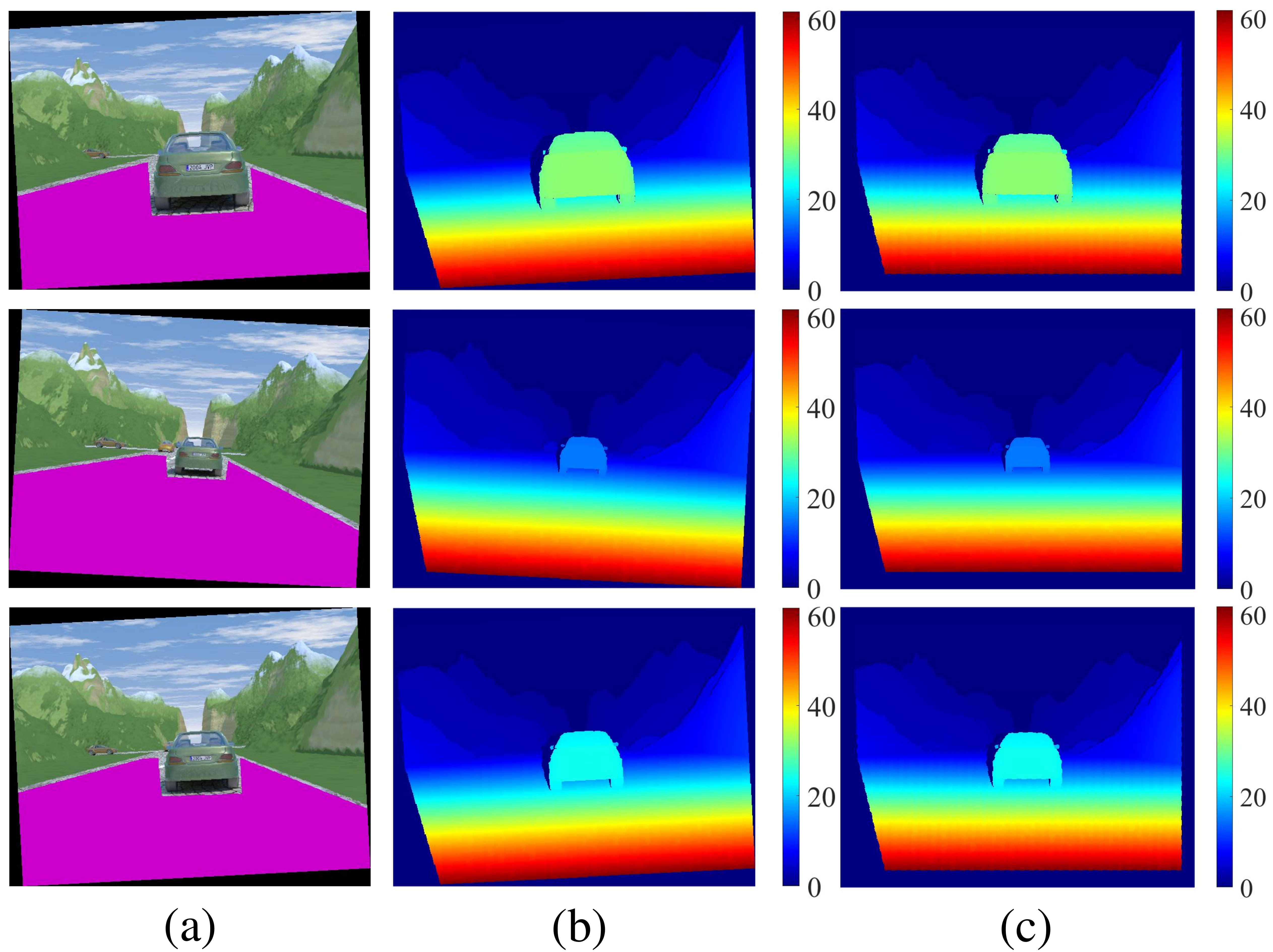}
		\caption{Examples of the roll angle estimation experiments; (a) reference images, the areas in magenta are the manually selected road regions; (b) original disparity maps; (c) disparity maps rotated around the estimated roll angles. }
				\vspace{-1em}
		\label{Figure eisats}
	\end{center}
\end{figure}

\begin{figure*}[!t]
	\begin{center}
		\centering
		\includegraphics[width=0.97\textwidth]{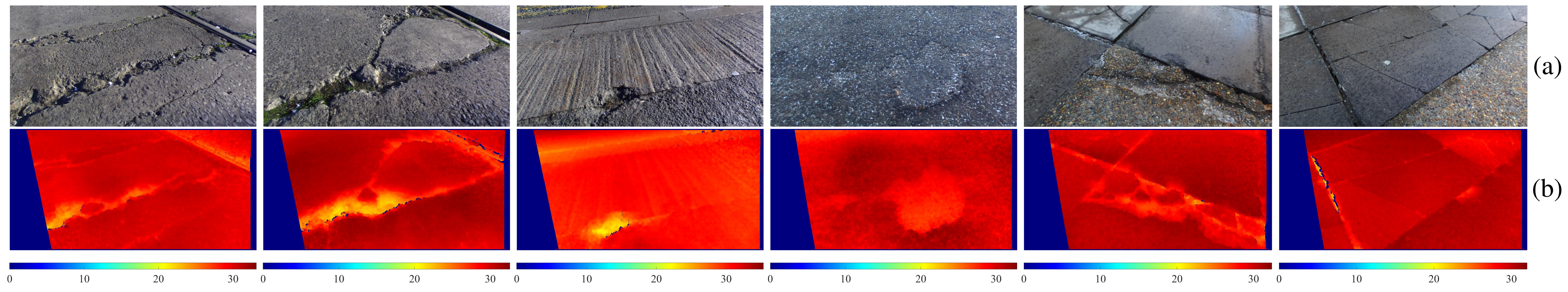}
		\caption{Examples of the disparity transformation experiments; (a) reference images; (b) transformed disparity maps. }
			\vspace{-1.em}
		\label{Figure rui_datasets_dt}
	\end{center}
\end{figure*}
\begin{figure*}[!t]
	\begin{center}
		\centering
		\includegraphics[width=0.98\textwidth]{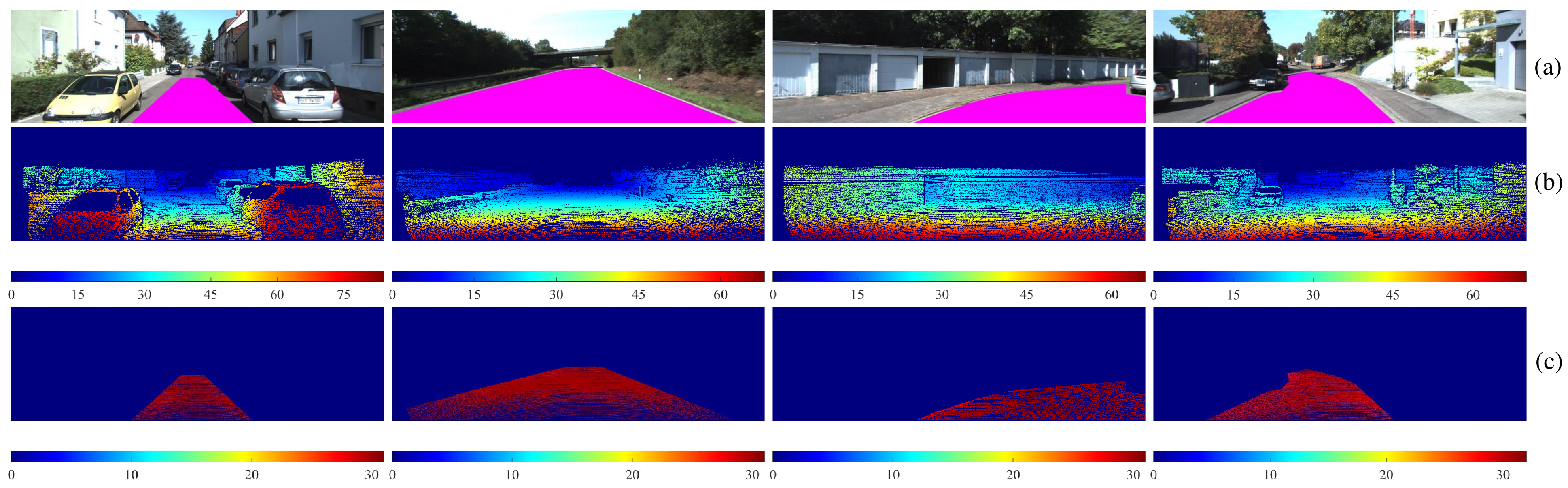}
		\caption{Disparity transformation experimental results of the KITTI stereo datasets;
			(a) reference images, where the areas in magenta are the manually selected road regions; (b) ground truth disparity maps; (c) transformed disparity maps. }
		\label{Figure disparity_transformation_kitti}
		\vspace{-1.5em}
	\end{center}
\end{figure*}

\subsubsection{Roll Angle Estimation}
In the experiments, we select a range of $\lambda^{(0)}$ and record the number of iterations that (\ref{eq.e_min_gd}) takes to converge to the minimum. It is shown that $\lambda^{(0)}=10$ is the optimum value when the threshold $\delta_\psi$ is set to $\frac{\pi}{1.8\times10^{6}}$ rad ($0.0001^\circ$).

Furthermore, a synthesised stereo dataset from EISATS\footnote{https://ccv.wordpress.fos.auckland.ac.nz/eisats/set-2/} \cite{Vaudrey2008, Wedel2008} is used to quantify the accuracy of the proposed roll angle estimation algorithm. The roll angle of each image in this dataset is perfectly zero. Therefore, we manually rotate the disparity maps around a given angle, and then estimate the roll angles from the rotated disparity maps. Examples of the roll angle estimation experiments are shown in Figure  \ref{Figure eisats}, where it can be observed that the effects due to image rotation are effectively corrected. When $\delta_\psi$ is set to  $\frac{\pi}{1.8\times10^{6}}$ rad, the average difference $\Delta \theta$ between the actual and estimated roll angles is approximately $0.012$ rad. The runtime of the proposed roll angle estimation on the Jetson TX2 GPU is approximately 7.842 ms.

\subsubsection{Disparity Transformation}
In \cite{Fan2018}, Fan \etal published three road datasets  containing various types of road damages, such as potholes and cracks. Therefore, we first use their datasets to qualitatively evaluate the performance of the proposed disparity transformation algorithm. Examples of the transformed disparity  maps are illustrated in Figure  \ref{Figure rui_datasets_dt}, where it can be observed that the disparities of the road surface have similar values, while their values differ greatly from those of the road damages. This fact enables the damaged road areas to be easily recognised from the transformed disparity maps. 

The KITTI stereo datasets are further utilised to evaluate the performance of disparity transformation. Examples of the KITTI stereo datasets are shown in Figure  \ref{Figure disparity_transformation_kitti}. To quantify the accuracy of the transformed disparities, we compute the standard deviation $\sigma_d$ of the transformed disparity values as follows:
\begin{equation}
\sigma_d=\sqrt{\frac{1}{m}\norm{{\mathbf{s}}-\frac{{\mathbf{s}}^\top \mathbf{1}_m}{m}}^2_2},
\label{eq.sigma_d}
\end{equation}
where $\mathbf{s}=[\ell'({\mathbf{p}}_0),\ell'({\mathbf{p}}_1),\dots,\ell'({\mathbf{p}}_{m-1})]^\top$  stores the transformed disparity values. The average $\sigma_d$ value of the KITTI stereo datasets is 
0.519 pixels. However, if the image rotation effects caused by the non-zero roll angle are not eliminated, the average $\sigma_d$ value becomes 0.861 pixels. The runtime of the disparity transformation on the Jetson TX2 GPU is around 1.541 ms.

\section{Conclusion and Future Work \label{sec.conclusion}}
This paper presented a robust dense stereo vision system embedded in a DJI Matrice 100 UAV for road condition assessment. The perspective transformation greatly improved the disparity accuracy and reduced the algorithm computational complexity, while the disparity transformation algorithm enabled the UAV to estimate roll angles from disparity maps. The damaged road areas became highly distinguishable in the transformed disparity maps, and this can provide new opportunities for UAV-based road damage inspection. The proposed system was implemented with CUDA on a Jetson TX2 GPU, and real-time performance was achieved. 

In the future, we plan to use the obtained disparity maps to estimate the flight trajectory of the UAV and reconstruct the 3D maps using the state-of-the-art simultaneous localisation and mapping (SLAM) algorithms. 

\section{Acknowledgment}
\label{sec.ack}
This work is supported by grants from  the Research Grants Council of the Hong Kong SAR Government, China (No. 11210017 and No. 21202816) awarded to Prof. Ming Liu. 
This work is also supported by grants from the Shenzhen Science, Technology and Innovation Commission, JCYJ20170818153518789, and National Natural Science Foundation of China (No. 61603376) awarded to Dr. Lujia Wang.

\balance
{\small
	\bibliographystyle{ieee}

}

\end{document}